%% file: main.tex
\documentclass{sig-alternate}
\usepackage{xcolor}
\usepackage{afterpage}
\usepackage{setspace}
\usepackage{framed}
\usepackage{tikz}
\usepackage{tikz-dependency}
\usepackage{soul}
\usepackage{graphicx}
\usepackage{times}
\usepackage{latexsym}
\usepackage{bbm}
\usepackage{amsmath}
\usepackage{multirow}
\usepackage{url}
\usepackage[FIGTOPCAP]{subfigure}
\usepackage{rotating}
\usepackage{pbox}
\usepackage{floatrow}

\definecolor{shadecolor}{rgb}{0.93,0.95,1.0}

\input{weld-defns-edited}

\newcommand{\instaread}{{\sc InstaRead}}

\newcommand\vardarrow[1]{\left\downarrow\rule{0pt}{#1}\right.}

\input{math}

\usepackage{etoolbox}
\makeatletter
\patchcmd{\maketitle}{\@copyrightspace}{}{}{}
\makeatother

\begin{document}


\title{Extreme Extraction: Only One Hour per Relation}

\numberofauthors{3}

\author{
\alignauthor
Raphael Hoffmann\\ 
	\affaddr{Department of Computer}\\
	\affaddr{\mbox{Science \& Engineering}}\\
	\affaddr{University of Washington}\\
	\affaddr{Seattle, WA, U.S.A.}\\
	\email{raphaelh@uw.edu}
\alignauthor
Luke Zettlemoyer \\
	\affaddr{Department of Computer}\\
	\affaddr{\mbox{Science \& Engineering}}\\
	\affaddr{University of Washington}\\
	\affaddr{Seattle, WA, U.S.A.}\\
	\email{lsz@cs.washington.edu}
\alignauthor
Daniel S. Weld \\
	\affaddr{Department of Computer}\\
	\affaddr{\mbox{Science \& Engineering}}\\
	\affaddr{University of Washington}\\
	\affaddr{Seattle, WA, U.S.A.}\\
	\email{weld@cs.washington.edu}
}

\date{2 February 2015}

\maketitle
\begin{abstract}
\input{abstract}

\end{abstract}




\vfill
\eject

\input{intro}

\input{problem}

\input{related}

\input{efficient}

\input{rulelanguage}

\input{visualizing}

\input{mixedinitiative}

\input{implementation}

\input{exp}

\input{discussion}




\vfill\eject

\nocite{hoffmann10}
\nocite{nakashole13}
\nocite{tokensregex2014}
\nocite{appelt98}
\bibliographystyle{abbrv}
\bibliography{main}

\input{appendix}

\end{document}

%% file: weld-defns-edited.tex
\newcommand{\comment}[1]{}

\newcommand{\bi}{\begin{list}{$\bullet$}{
    \setlength{\leftmargin}{1.5 em}
    \setlength{\itemsep}{0 pt}
    \setlength{\topsep}{3 pt}
    \setlength{\parsep}{3 pt}
    \setlength{\partopsep}{0 pt}
    \setlength{\labelwidth}{1 em}
    \setlength{\labelsep}{0.5 em}
    \setlength{\parskip}{0cm}  }}
\newcommand{\ei}{\end{list}}

\newcommand{\BE}{\begin{enumerate}}
\newcommand{\EE}{\end{enumerate}}

\newcommand{\initab}{                           
\begin{tabbing}
XXX \= XXXX \= \kill
}
\newcommand{\begpub}{
\begin{quotation}
\noindent
}

\newcommand{\finpub}{
\end{quotation}
}

%% file: math.tex
\DeclareMathOperator{\prepof}{\sf prep-of}
\DeclareMathOperator{\nsubjpass}{\sf nsubjpass}
\DeclareMathOperator{\killOfNom}{\sf killOfNom}
\DeclareMathOperator{\token}{\sf token}
\DeclareMathOperator{\nn}{\sf nn}
\DeclareMathOperator{\killerRole}{\sf killerRole}
\DeclareMathOperator{\killingBNF}{\sf killingBNF}
\DeclareMathOperator{\killingBInf}{\sf killingBInf}
\DeclareMathOperator{\dobj}{\sf dobj}
\DeclareMathOperator{\killed}{\sf killed}
\DeclareMathOperator{\person}{\sf person}
\DeclareMathOperator{\prepby}{\sf prep-by}
\DeclareMathOperator{\poss}{\sf poss}
\DeclareMathOperator{\appos}{\sf appos}
\DeclareMathOperator{\actInd}{\sf actInd}
\DeclareMathOperator{\passInd}{\sf passInd}
\DeclareMathOperator{\dep}{\sf dep}
\DeclareMathOperator{\rcmod}{\sf rcmod}
\DeclareMathOperator{\agent}{\sf agent}
\DeclareMathOperator{\infmod}{\sf infmod}
\DeclareMathOperator{\partmod}{\sf partmod}
\DeclareMathOperator{\xsubj}{\sf xsubj}
\DeclareMathOperator{\nsubj}{\sf nsubj}
\DeclareMathOperator{\xcomp}{\sf xcomp}
\DeclareMathOperator{\prepcof}{\sf prepc-of}
\DeclareMathOperator{\prepcfor}{\sf prepc-for}
\DeclareMathOperator{\prepcwith}{\sf prepc-with}
\DeclareMathOperator{\prt}{\sf prt}
\DeclareMathOperator{\prepin}{\sf prep-in}
\DeclareMathOperator{\prepto}{\sf prep-to}
\DeclareMathOperator{\prepwith}{\sf prep-with}
\DeclareMathOperator{\prepfor}{\sf prep-for}
\DeclareMathOperator{\prepcto}{\sf prepc-to}

%% file: abstract.tex
Information Extraction (IE) aims to automatically generate a large
knowledge base from natural language text, but progress remains slow.
Supervised learning requires copious human annotation, while 
unsupervised and weakly supervised approaches do not deliver
competitive accuracy. As a result, most fielded applications of IE, as
well as the leading TAC-KBP systems, rely on significant amounts of
manual engineering. Even ``Extreme'' methods, such as those reported
in Freedman et al.~\shortcite{freedman11}, require about 10 hours of
expert labor per relation.

This paper shows how to reduce that effort by an order of magnitude.
We present a novel system, {\sc InstaRead}, that streamlines authoring
with an ensemble of methods: 1) encoding extraction rules in an expressive
and compositional representation, 2) guiding the user to promising
rules based on corpus statistics and mined resources, and 3) introducing a
new interactive development cycle that provides immediate feedback ---
even on large datasets. Experiments show that experts can create
quality extractors in under an hour and even NLP novices can author
good extractors. These extractors equal or outperform ones obtained
by comparably supervised and state-of-the-art distantly supervised approaches.

%% file: intro.tex
\section{Introduction}

Information Extraction (IE), the process of distilling semantic
relations from natural language text, continues to gain
attention. If applied to the Web, such IE systems have the potential
to create a large-scale knowledge base which would benefit
important tasks such as question answering and summarization.

Applying information extraction to many relations, however, remains a
challenge. One popular approach is supervised learning of
relation-specific extractors, but these methods are limited by the
availability of training data and are thus not scalable. Unsupervised
and weakly supervised methods have been proposed, but are not
sufficiently accurate. Many successful applications of IE therefore
continue to rely on significant amounts of manual engineering. For
example, the best performing systems of the TAC-KBP slot filling
challenge make central use of manually created
rules~\cite{min12,sun11}.

In response, Freedman et al.~\shortcite{freedman11} proposed {\em
  Extreme Extraction}, a combination of techniques which enabled
experts to develop five slot-filling extractors in 50 hours, starting
with just 20 examples per slot type. 
These extractors outperformed ones learned with manual supervision 
and also required less 
effort, when 
data labeling costs were included.

In this work, we seek to dramatically {\em streamline} the process of
extractor engineering, while handling the more general task of binary
relations, $r(a, b)$, where both arguments are free.  Our
goal is to enable researchers to create a high-quality relation
extractor in under one hour, using no prelabeled data. To achieve this
goal, we propose an extractor development tool, \instaread,
which defines a user-system interaction based on three key
properties. First, experts can write {\em compositional} rules in an expressive logical
notation.  Second, the system guides the expert to promising rules,
for example through a bootstrap rule induction algorithm which
leverages the distribution of the data. Third, rules can be tested
instantly even on relatively large datasets.

This paper makes the following contributions:

\bi
\item We present \instaread, an integrated ensemble of methods for rapid
extractor construction.
\item We show how these components can be implemented to
enable {\em real-time} interactivity over millions of documents.
\item We evaluate \instaread\ empirically, showing 1) an expert user can
  quickly create high precision rules with large recall. \footnote{
  All rule sets developed as part of this work, the training data produced by
  odesk workers, and the output extractions from each system are available
  upon request.
} 
  that greatly outperform
  comparably supervised and state-of-the-art distantly supervised approaches
  and require one tenth the manual effort of Freedman's~\shortcite{freedman11} approach. We
  also present 2) 
  the cumulative gains due to different \instaread\ features, as well as 3) an
  error analysis indicating that more than half of extractor mistakes stem from
  problems during preprocessing (e.g., parsing or NER). 
We further show that 
4) even NLP novices can use \instaread\ to create
  quality extractors for many relations. 

\ei

%% file: problem.tex
\section{Problem Definition}
\label{sec:problem}

Engineering competitive information extractors
often involves the development of a carefully selected set of rules.
The rules are then used in a number of ways, for example, to
create deterministic rule-based extractors~\cite{min12,sun11}, or as features or constraints
in learning-based systems~\cite{poon07,ganchev10}.
Typically, however, the development of rules is an
iterative process of refinement that involves (1) analyzing a development corpus of text
for variations of relation mentions, (2) creating hypotheses for how these
can be generalized, (3) formulating these hypotheses in a rule language,
and (4) testing the rules on the corpus.  

Unfortunately, each of the steps in this cycle can be very time intensive.
For example, when analyzing a corpus an expert may spend much time searching for relevant sentences.
When creating hypotheses, an expert may not foresee
possible over- or under-generalizations. An expert's intended generalization
may also not be directly expressible in a rule language, and testing may be
computationally intensive in which case the expert is unable to obtain immediate feedback.

Our goal in this work is to develop and compare techniques which accelerate this cycle, 
so that engineering a competitive extractor requires less than an hour of expert time.

%% file: related.tex
\section{Prior Work}
\label{sec:instarelated}

Freedman et al.~\shortcite{freedman11}'s landmark work on `Extreme Extraction' first articulated the important challenge of investigating the development of information extractors within a limited amount of time. Their methods, which allowed an expert to create a question answering system for a new domain in one week, used orders of magnitude less human engineering than the norm for ACE\footnote{\tiny{\url{http://www.itl.nist.gov/iad/mig/tests/ace/}}}, TAC-KBP and MUC\footnote{\tiny{\url{http://www-nlpir.nist.gov/related_projects/muc}}} competitions. 
Key to this improvement was a hybrid approach that combined 
a bootstrap learning algorithm and manual rule writing. Freedman et al. 
showed that this combination yields higher recall and F1 compared to approaches that used
only bootstrap learning or manual rule writing, but not both.

Freedman et al.'s task is related but different from our task; some of these differences make it harder and others easier. In particular, Freedman et al. sought to extract relations $R_i(arg_1, arg_2)$, in which one
of the arguments was fixed. A small amount of training data was assumed for each relation,
and the task included adaptation of a named entity recognizer and coreference resolution system.

With \instaread\ we propose a combination of different, complementary techniques to those of Freedman et al., that focus on streamlining rule authoring. One of these techniques leverages a refined and very effective bootstrap algorithm that keeps the user in the loop, whereas Feedman et al.'s bootstrap learner ran autonomously without user interaction.


A large amount of other work has looked
at bootstrapping extractors from a set of seed examples. 
Carlson et al.'s NELL~\shortcite{carlson09b} performs
coupled semi-supervised learning to extract a broad set
of instances, relations, and classes from a Web-scale corpus.
Two of the four relations we report results for in Section~\ref{sec:exp} are covered by NELL, yielding
174 (attendedSchool) and 977 (married) instances after 772 iterations.
To avoid a decline in precision (on average 57\% after 66 iterations), NELL relies on periodic human supervision of 5 minutes per relation every 10 iterations.
Other systems leverage large knowledge bases for supervision.
PROSPERA~\cite{nakashole11} uses MaxSat-based
constraint reasoning and improves
on the results, but requires that relation arguments be
contained in the YAGO knowledge base~\cite{suchanek08}.
Only one of our four relations is extracted by PROSPERA (attendedSchool), yielding 1,371
instances at 78\% precision. 26,280 instances of this relation from YAGO
were used for supervision. DeepDive~\cite{niu12, niu12x, re14} scales a Markov logic program to
the same corpus and uses Freebase for supervision. It reaches an F1 score of 31\% on 
the TAC-KBP relation extraction benchmark.
MIML-RE~\cite{mihai12} and MultiR~\cite{hoffmann11}, also apply distant supervision from a knowledge base but add global constraints to relax the assumption that every matching sentence expresses a fact in the knowledge base. 
Except for the latter two systems, which we compared to, none of the above systems is publicly released, 
making a direct comparison impossible.
In general, all of the above approaches suffer from relatively low recall and combat semantic
drift by relying on redundancy, global constraints, large knowledge bases, or validative feedback.
\nocite{riedel10}

While above approaches attempt to avoid manual input altogether, other
approaches try to make manual input more effective. These include compositional
pattern languages for specifying rule-based extractors, such as CPSL~\cite{appelt98}
and TokensRegex~\cite{tokensregex2014}. Rule-based extraction has also been scaled to
larger datasets by applying query optimization~\cite{KrishnamurthyLRRVZ08,ChiticariuKLRRV10}.
Unlike our work, this line of research does not evaluate the effectiveness of these
languages with users, in terms of development time and extraction quality.
Another approach to human input is active learning. Miller et al.~\shortcite{miller04} learn a 
Perceptron model for named-entity extraction; unlabeled
examples are ranked by difference in perceptron score.
Riloff~\shortcite{Riloff96} proposes an approach to
named-entity extraction, which requires users to first
classify documents by domain, and then generates and 
ranks candidate extraction patterns.
Active learning has also been studied in more general
contexts, for learning probabilistic models with labeled
instances~\cite{thompson99} or labeled features~\cite{druck09}.
This work differs from approaches based on active learning 
in at
least two ways. First, they have not been evaluated on
relation extraction tasks. Second, and more importantly,
their general approach is to consider a particular type of 
feedback and then develop algorithms for learning more
accurately from such feedback. In contrast, our approach
is not to compare algorithms, but to compare different types
of manual feedback.

%% file: efficient.tex
\section{Overview of InstaRead}
\label{sec:instaoverview}

To evaluate techniques for accelerating the development process of rule-based extractors,
we developed \instaread, an interactive extractor development tool. \instaread\ is designed
to address the inefficiencies identified in Section~\ref{sec:problem} with the combination
of an expressive rule language, data-driven guidance to new rules, and instant rule execution.

We next present an example of an expert interacting with the
system, and then in the following sections show how \instaread\ enables its three key properties.

\subsection{Example}
\label{sec:example}

\newcommand{\cfbox}[2]{%
    \colorlet{currentcolor}{.}%
    {\color{#1}%
    \fbox{\color{currentcolor}#2}}%
}

{%
\setlength{\fboxsep}{0pt}%
\setlength{\fboxrule}{0.5pt}%

\definecolor{light-gray}{gray}{0.8}

\renewcommand{\subfigtopskip}{0pt}
\renewcommand{\subfigbottomskip}{0pt}
\renewcommand{\subfigcapskip}{-10pt}
\renewcommand{\subfigcapmargin}{0pt}

\begin{figure*}[htb]
\centering
\subfigure{
  \centering
  \cfbox{light-gray}{\includegraphics[width=6in, trim=0cm 6.9cm 0cm 0cm, clip=true,natwidth=571,natheight=481]{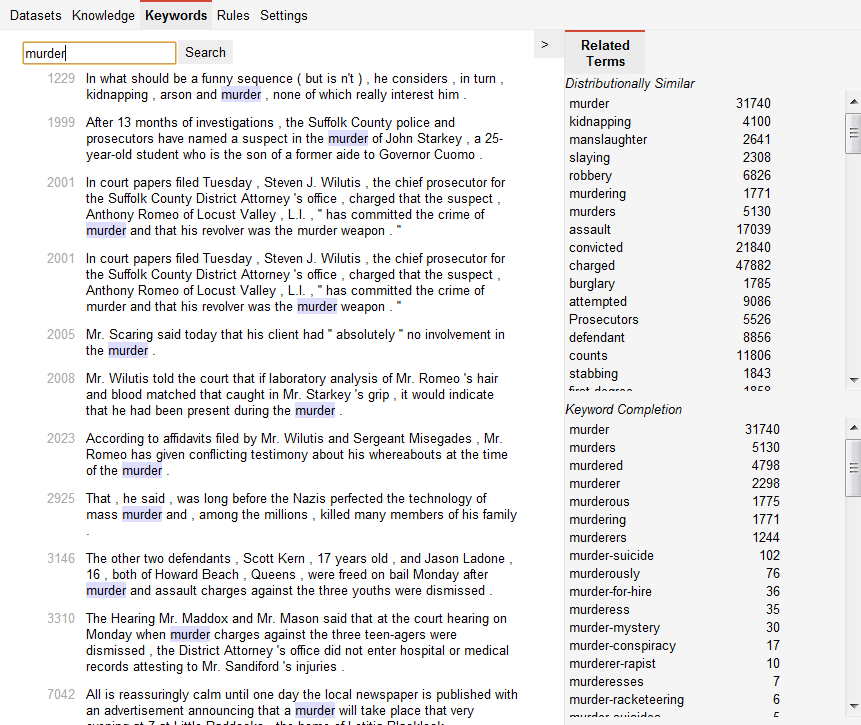}}
} \\
\vspace{.2in}
\subfigure{
  \centering
  \cfbox{light-gray}{\includegraphics[width=6in, trim=0cm 6.9cm 0cm 0cm, clip=true,natwidth=854,natheight=726]{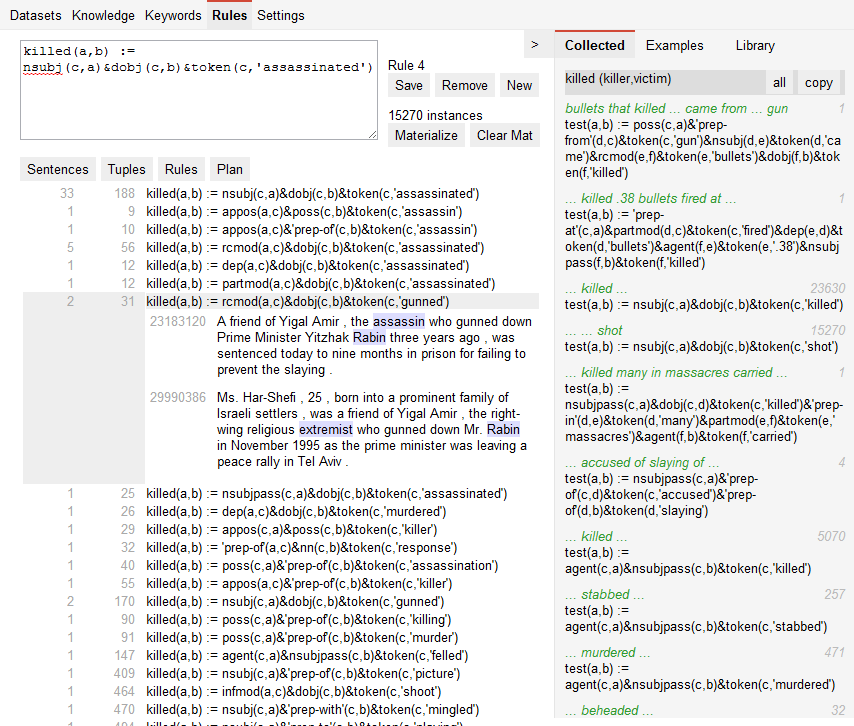}}
}
\caption{Selected interactions (Section~\ref{sec:instaoverview}).
\instaread\ guides users to sentences and rules by distributionally-similar words (top), and
bootstrap pattern learning (bottom).
}
\label{fig:insta}
 
\end{figure*}

\renewcommand{\subfigtopskip}{10pt}
\renewcommand{\subfigbottomskip}{10pt}
\renewcommand{\subfigcapskip}{10pt}
\renewcommand{\subfigcapmargin}{10pt}

}%

Anna wants to create an extractor for the {\sf killed(killer,victim)} relation. After selecting a development text corpus she proceeds as follows.

\begin{itemize}
\item To find example sentences, Anna searches for sentences containing keyword `killed' (Figure~\ref{fig:insta} a). \instaread\ suggests to also consider distributionally similar keywords such as `murder' or `assassin'. Within seconds Anna obtains many relevant examples.

\item Anna compares examples, investigating their syntactic structure obtained by a parser, and encodes an extraction pattern as a rule: $\sf killed(a,b) \Leftarrow nsubj(c,a) \& dobj(c,b) \& $\\$\sf token(c, 'murdered')$. The system offers to automatically generalize that rule, so that it also covers the passive form as well as all tenses.

\item Anna now has a working extractor which she would like to refine. \instaread's bootstrapping method presents her a ranked list of new candidate rules based on the extractions of her existing rule set. Anna inspects matches of the suggested rules and selects several (Figure~\ref{fig:insta} b).

\item Looking at the rules collected so far, Anna notices that many are similar, differing only in the verb that was used. She decides to refactor her rule set, so that one rule first identifies relevant verbs, and others  syntactic structure. Her rule set is now more compact and generalizes better.

\end{itemize}

%% file: rulelanguage.tex
\section{Condition-Action Rules in Logic}
\label{sec:rulelanguage}

A rule language should be both simple and expressive, so that the interaction
with the system is quick and direct. To fulfill this requirement, \instaread\
uses condition-action rules expressed in first-order logic, combined
with a broad and expandable set of built-in logical predicates. For tractability, \instaread\ requires that rules translate into {\em safe} domain-relational calculus~\cite{Ullman88}.

Although such rules could be used to generate statistical models~\cite{2009Domingos}, we
currently assume that all rules are deterministic and are executed in a defined order, and leave 
the integration with learning-based techniques as future work.

\begin{figure}[ht]
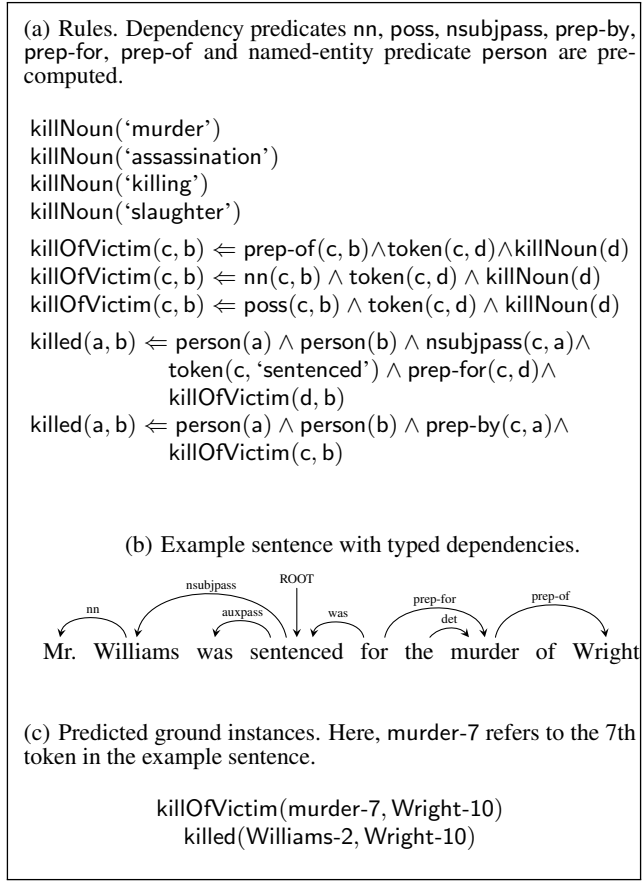

\setlength{\FrameSep}{6pt}
\begin{framed}
\subfigure[Rules. Dependency predicates $\sf nn$, $\sf poss$, $\sf nsubjpass$, 
$\sf prep\textrm{-}by$, $\sf prep\textrm{-}for$, $\sf prep\textrm{-}of$ and named-entity predicate 
$\sf person$ are pre-computed.]{
\centering
\begin{minipage}{\linewidth}
\vspace{0.1in}
$\sf killNoun(\textrm`murder\textrm')$\\
$\sf killNoun(\textrm`assassination\textrm')$ \\
$\sf killNoun(\textrm`killing\textrm')$ \\
$\sf killNoun(\textrm`slaughter\textrm')$ \par
\vspace{.05in}
\noindent
$\sf killOfVictim(c,b) \Leftarrow prep\textrm{-}of(c,b) \wedge token(c, d) \wedge killNoun(d)$ \\
$\sf killOfVictim(c,b) \Leftarrow nn(c,b) \wedge token(c, d) \wedge killNoun(d)$ \\
$\sf killOfVictim(c,b) \Leftarrow poss(c,b) \wedge token(c,d) \wedge killNoun(d)$ \par
\vspace{.05in}
\noindent
$\sf killed(a,b) \Leftarrow person(a) \wedge person(b) \wedge nsubjpass(c,a) \wedge$ \\
\phantom{$\sf killed(a,b) \Leftarrow $}$\sf token(c, \textrm`sentenced\textrm') \wedge prep\textrm{-}for(c,d) \wedge  $\\ 
\phantom{$\sf killed(a,b) \Leftarrow $}$\sf killOfVictim(d,b)$ \\
$\sf killed(a,b) \Leftarrow person(a) \wedge person(b) \wedge prep\textrm{-}by(c,a) \wedge$ \\
\phantom{$\sf killed(a,b) \Leftarrow $}$\sf killOfVictim(c,b) $ \\
\end{minipage}
}\par
\vspace{0.1in}
\subfigure[Example sentence with typed dependencies.]{
\centering
\vspace{0.1in}
\begin{dependency}[theme = simple]
   \begin{deptext}[column sep=0.25em]
      Mr. \& Williams \& was \& sentenced \& for \& the \& murder \& of \& Wright \& . \\
   \end{deptext}
   \deproot[edge unit distance=1.5ex]{4}{ROOT}
   \depedge{2}{1}{nn}
   \depedge{4}{2}{nsubjpass}
   \depedge[edge start x offset=-6pt]{4}{3}{auxpass}
   \depedge[edge end x offset=+6pt]{5}{4}{was}
   \depedge{5}{7}{prep-for}
   \depedge[arc angle=50, edge end x offset=-6pt]{6}{7}{det}
   \depedge{7}{9}{prep-of}
\end{dependency}
}
\par
\vspace{.1in}
\subfigure[Predicted ground instances. Here, $\sf murder\textrm{-}7$ refers to the 7th token in the example sentence.]{
\begin{minipage}{\linewidth}
\vspace{0.1in}
\centering
$\sf killOfVictim(murder\textrm{-}7, Wright\textrm{-}10)$\\
$\sf killed(Williams\textrm{-}2, Wright\textrm{-}10)$\\
\end{minipage}
}
\end{framed}
\caption{Example rule set executed on sentence.}
\label{fig:setofrules}
\end{figure}

Figure~\ref{fig:setofrules} presents an example rule set and how it is applied to a sentence. The predicates used in this example have arguments that range over tokens and token positions. Rules are used to define predicate {\sf killOfVictim} and that predicate then gets re-used in other rules to define predicate {\sf killed}. 
We call this ability \instaread's {\em Composition} feature.

To increase the expressiveness of the language, \instaread\ implements around one hundred
built-in predicates, such as {\sf tokenBefore} and {\sf isCapitalized}. In addition, it
makes available predicates that encode the output of (currently) four NLP systems,
including a phrase structure parser~\cite{CharniakJ05}, a typed dependency extractor~\cite{marneffe06}\footnote{\instaread\ uses collapsed dependencies with propagation of conjunct dependencies.}, a coreference resolution system~\cite{RaghunathanLRCSJM10}, and a named-entity tagger~\cite{finkel-acl05}. This allows users to write rules which simultaneously use parse, coreference, and entity type information.


The rules in Figure~\ref{fig:setofrules} are all Horn clauses, but \instaread\ also supports disjunction ($\vee$), negation ($\neg$), and existential ($\exists$) and universial ($\forall$) quantification. While one does not often need these operators, they are sometimes convenient for specific lexical ambiguities. For example, in our evaluation discussed in section~\ref{sec:exp}, one user of \instaread\ created the following rule to extract instances of the {\sf founded(person,organization)} relation:

\vspace{-.1in}
{\scriptsize
\begin{align*}
\sf founded(a,b) \Leftarrow & \sf nsubj(c,a) \wedge dobj(c,b) \wedge token(c, \textrm`built\textrm') \wedge \\
 & \sf person(a) \wedge organization(b)
\end{align*}
}%
This rule was designed to match sentences such as: `Michael Dell built his first company in a dorm-room.' However, this rule also incorrectly matches a number of other sentences such as:
`Mr. Harris built Dell into a formidable competitor to IBM.'
While `building an organization' typically implies a {\sf founded} relation, `building an organization {\em into} something' does not. This distinction can be captured in our rule by adding the conjunct $\sf \neg(\exists d : prep\textrm{-}into(c,d))$.

%% file: visualizing.tex
\section{Guiding Experts to Effective \\Rules}
\label{sec:accelerators}

While the rule language is important for developing extractors, many hours of testing
on early prototypes of \instaread\ showed that it is not sufficient for an effective
interaction. With a growing number of rules, users find it increasingly difficult to identify rules
for refinement. More importantly, users don't know where to focus their attention
when trying to find effective rules to add.

\begin{figure}[ht]
\centering
\includegraphics[width=3.3in]{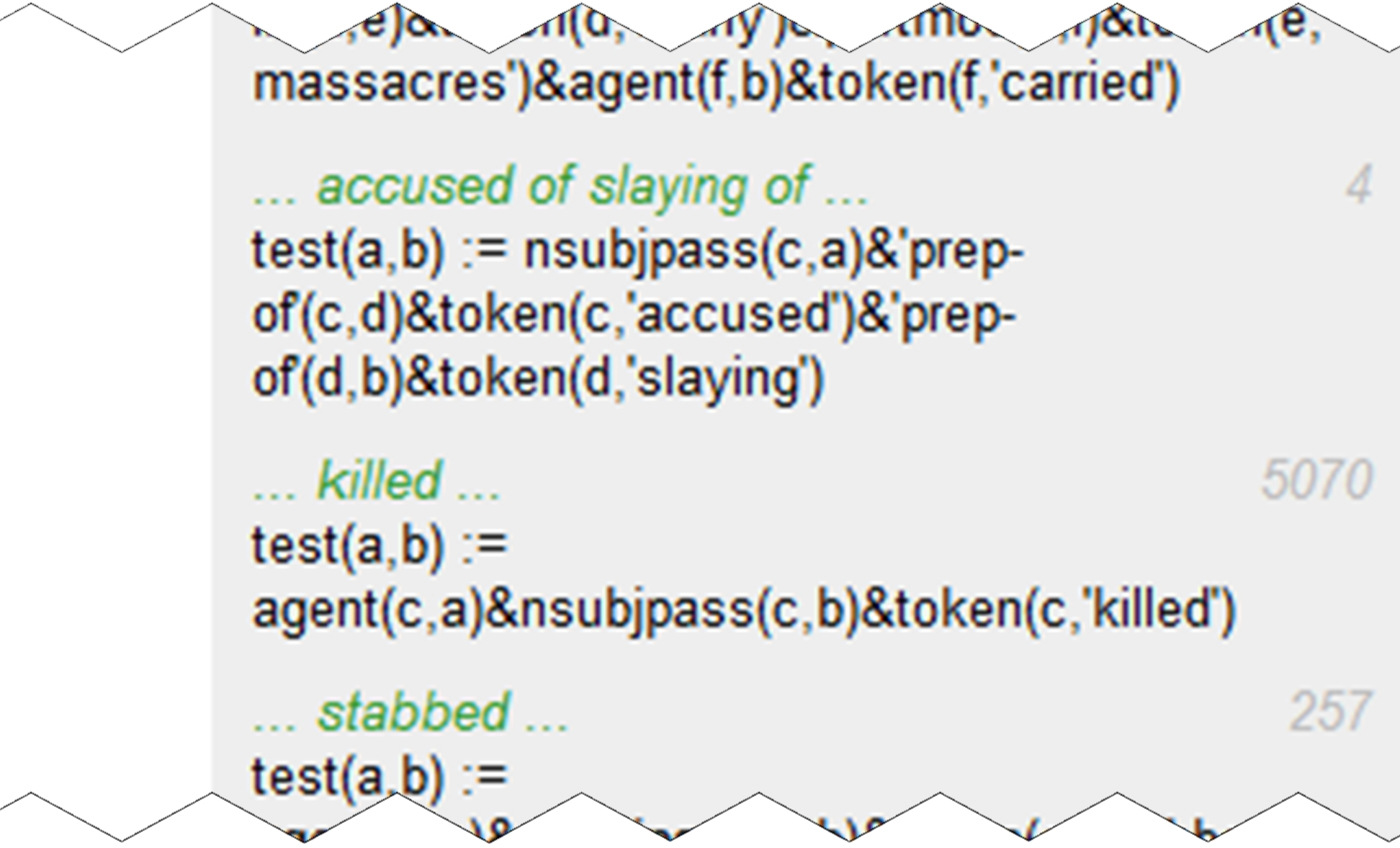}
\caption{\instaread\ shows automatically generated comments and number of extractions together
with each rule, allowing users to see (approximate) meaning and relevance without needing to read logical expressions.}
\label{fig:instavis2}
\end{figure}

\paragraph{Feedback} Our small example in Figure~\ref{fig:setofrules} already 
demonstrates the problem: What exactly does each rule do? How much data does each rule affect? 
In early testing, we noticed that users would often write
short comments for each rule, consisting of the surface tokens matched. We therefore 
designed a technique to automatically generate such comments (depicted in Figure~\ref{fig:instavis2})
by retrieving matched
sentences, identifying sentence tokens that were explicitly referenced by one of the 
predicates, and concatenating the tokens in the order that they appear in a sentence. Included
are `\ldots' placeholders for the arguments of the rule's target predicate.
Figure~\ref{fig:instavis2} also shows how \instaread\ displays the number of matches together with each rule, eg. 257 for `\ldots stabbed \ldots', helping users quickly judge the importance of a rule. Although one may also be interested in precision, that cannot be obtained
without annotated data.
\instaread\ also includes visualizations for dependency trees, parse trees, and coreference clusters.
Such visualizations do not always convey all information encoded in the logical representation,
but convey (approximate) meaning or relevance quickly.

%% file: mixedinitiative.tex
\paragraph{Bootstrap Rule Induction}
How does an expert know what rules to write? Coming up with good candidates
is surprisingly difficult. One approach is automatic rule suggestions based
on statistics. This can be done, for example, using 
a semi-supervised bootstrap pattern learning algorithm. Freedman et al.~\shortcite{freedman11} 
applied such an algorithm, too, but found that it was not competitive with
manual pattern writing, especially with regards to recall.
\instaread's bootstrap algorithm therefore makes several changes:
First, it instantly returns ranked bootstrap results over a large corpus. Second, it takes into account
coreference information to expand recall (similarly to Gabbard et al.~\shortcite{gabbard11}).
Third, it puts the user into the loop, allowing her to select appropriate
rules after each iteration.

In particular, \instaread's bootstrap technique takes as input a binary
relation predicate $r(a,b)$ together with a set of rules $R$ defining 
instances of $r$. The output
is a ranked list of candidate rules $S$. The algorithm works by first
identifying mentions of $r$ using the existing rules $R$ and generating the pairs 
$(a_s,b_s)$ of argument surface strings of these mentions. This set
of pairs is then matched to the entire corpus, retrieving
all sentences containing both strings. Similar to DIRT~\cite{lin01}, \instaread\ then generates
syntactic-lexical extraction patterns from these matches. 
Loosely following Mintz et al.~\cite{mintz09}, the system finds a path
of syntactic dependencies connecting the matched surface strings, and then
creates a rule that is a conjunction of syntactic dependencies and lexical
constraints on that path, as well as entity type constraints (if activated by user). 
For examples, refer to Figure~\ref{fig:insta}.

Rule suggestions, $S$, are sorted by two scores: pointwise mutual information of 
the suggested rule with the original rule set $R$, and number of extractions of
suggested rule. The latter may show more irrelevant rules on top, but the 
relevant ones among them have many extractions often reducing overall effort.
Users can switch between the sort orders.
 
\paragraph{Word-level Distributional Similarity}
Although our enhancements to the bootstrap approach may increase recall, recall
is still limited since bootstrap requires that the same tuples appear multiple times in the corpus.
To help experts find additional relation mentions, \instaread\ therefore also includes 
another shallow technique: keyword search combined with keyword suggestions.

Keywords are suggested based on distributional similarity to a seed keyword.
For example, the seed `murdered' returns `assassinated', `slayed', `shot',
and more. Specifically, each word $w$ in the text corpus is represented as a vector
of weighted words $v$ co-occurring in sentences with $w$. The similarity of
two words $w_1, w_2$ is then defined as the cosine similarity of their vector
representations. An additional list of keyword suggestions shows keywords
which contain the seed keyword as prefix. Suggested keywords are always
displayed together with their number of occurrences in the corpus to guide
users to the most relevant keywords.

Although this keyword-based approach may be effective in finding relevant
sentences, early experiments have shown that a long time is spent to writing
extraction rules based on those sentences. We therefore added a simple interface
feature: Experts could click on words to indicate relation arguments, and the
system will generate rule candidates using our bootstrap generation algorithm.

\paragraph{Core Linguistic Rules}

The final problem we address is the fact that a relation extractor
typically needs a large number of rules that are not specific to the relation.
For example, there exist many syntactic variations that follow common linguistic
patterns. To reduce effort, we seek to populate the system with
such general rules right from the start.

In a first step, we encoded a set of grammatical rules: Given a verb base form,
\instaread\ can generate rules encoding syntactic-lexical patterns for 182
combinations of tense, voice, and person. For example, given subject X, object Y
and verb `kill', the system generates rules to capture phrases such as
`Y was killed by X', `X regretted killing Y', `X would later kill Y'. To avoid inaccuracies 
from using a stemmer, \instaread\ includes inflection rules and a corpus of inflections
for 16851 verbs mined from Wiktionary. This grammatical background knowledge is provided to the user
through a set of additional built-in predicates.

%% file: implementation.tex
\section{Efficient Rule Evaluation}
\label{sec:implementation}

To enable its interactivity, \instaread\ must evaluate rules and
guide users to effective rules very quickly, 
even with compositional rules and large datasets. 

\instaread\ is built on top of an RDBMS. Variables in its logical
expressions are assigned a data type that can be 
$\sf Pos$ (token position), $\sf Span$ (token span), $\sf Int$ (integer), 
$\sf Str$ (string), or $\sf Ref$ (reference). Each of these data types is internally
mapped to a composite SQL data type. For example, token spans are mapped
to the SQL types $\sf integer, byte, byte$, where the first is used to identify a
sentence and the others indicate start and end positions within a sentence.
Predicates are either {\em extensional} or {\em intensional}. Extensional
ones materialize instances in relational tables, while intensional ones
are defined by (partial) SQL queries. 
An example of an extensional predicate is our $\sf killed(a,b)$ extractor, which stores
the result set of the extraction rules depicted in Figure~\ref{fig:setofrules}. 
An example of an intensional predicate is $\sf str2span(s,t)$ which returns
all mentions of a multi-word string using an inverted index. For details on
how this predicate gets translated into SQL, see Figure~\ref{fig:instaintext} in the appendix.
The key component of \instaread's
implementation is its translation of logical rules into SQL queries.
The system first parses logical rules into an abstract syntax tree (AST).
To ensure that the rules do not yield infinite result sets and can be translated
into SQL, it checks for {\em safety}~\cite{Ullman88}. It then infers variable types and links predicates, 
then translates into an AST of tuple relational calculus, and eventually SQL, following the
algorithms described in ~\cite{Ullman88}. For an example translation, see Appendix A.

For performance, \instaread\ creates a BTree index for each column of
an extensional predicate. Built-in predicates (which tend to contain more
instances, e.g. all syntactic dependencies), also use multi-column indices.
A variety of information is pre-computed on a Hadoop cluster, including 
phrase structure trees, dependencies, coreference clusters, named-entities, 
rule candidates for bootstrapping, and distributionally similar words. 

This large number of indices and pre-computed information is important
because \instaread\ does not constrain the set of queries and most queries
touch the entire text corpus. It also allows each iteration of bootstrapping
to be performed by a single SQL query.
Across all of the experiments the median query execution time was 74ms.
Achieving such interactivity is crucial for quickly building accurate extractors.

%% file: exp.tex
\section{Experiments}
\label{sec:exp}

In our evaluation, we measured if \instaread's features enable
an expert to create quality extractors in less than one hour, 
and which of the features contribute most to reducing effort.
We also report on an error analysis 
to get insights into potential future improvements.
Finally, we report early results of a follow-up experiment, in which
we evaluated \instaread's usability among
engineers without NLP background.

\subsection{Experimental Setup}

We evaluated the performance of an expert in a controlled experiment, in which the expert user
was given one hour of time per relation to develop four relation
extractors. Besides descriptions of the four relations and a corpus of (unlabeled) news articles,
which was loaded into \instaread, no other resources were provided. Our expert 
was familiar with \instaread\ and NLP in general, but had no 
experience with the relations tested.
All user and system actions were logged together with their timestamps. 

We were interested in determining the effectiveness of four of \instaread's
features: bootstrap rule induction {\em (Bootstrap)}, word-level distributional similarity {\em (WordSim)},
core linguistic rules {\em (Linguistics)}, and the power of rule (de-)composition {\em (Composition)}. 
In order to more easily measure the impact of each of these features, our
 user was required to use only one at a time and switch to the next at given 
time intervals (Figure~\ref{fig:comparetools}).

\begin{table*}[htb]
\centering
\begin{tabular}{lrrrrrrrr}
 &   \multicolumn{2}{c}{\footnotesize{\sf attendedSchool}} & \multicolumn{2}{c}{\footnotesize{\sf founded}} & \multicolumn{2}{c}{\footnotesize{\sf killed }} & \multicolumn{2}{c}{\footnotesize{\sf married }} \\
\hline
 & Pr & \#e & Pr & \#e & Pr & \#e & Pr & \#e \\
InstaRead & {\bf 100\%} & {\bf 52,338} & 91\% & {\bf 20,733} &  {\bf 90\%} & {\bf 4,728} & 90\%  & {\bf 63,742} \\
MIML-RE & 9\% & 14,960 & 28\% & 14,960 & N/A & 0* & \bf{93\%} & 9,900 \\
MultiR & 26\% & 18,480 & 38\% & 10,340 & N/A & 0* & 51\% & 24,200 \\
SUP & 12\% & 25,196 & {\bf 100\%} & 2,255 & N/A & 0 & 44\% & 7,867 \\
\hline
\end{tabular}
\caption{Precision (Pr) and number of extractions (\#e) for the NYTimes test dataset. *Cases where extraction could not be performed because no target database could be found that contained examples required for distant supervision.
}
\label{tab:extrnyt}
\end{table*}

\paragraph{Baselines} We compare the performance of the extractors created with our proposed system \instaread\ to
three baselines.

{\bf MIML-RE}~\cite{mihai12} and {\bf MultiR}~\cite{hoffmann11} are two state-of-the-art systems for learning relation extractors by distant supervision from a database. As a database we use the instances of the relations contained in Freebase~\cite{bollacker08}. Negative examples are generated from random pairs of entity mentions in a sentence.\footnote{We added negative examples at a ratio of 50:1 to positives. Increasing this ratio increases precision but reduces the number of extractions, while decreasing has the opposite effect. We found that this setting provided a better trade-off than the default used by these distantly supervised systems on the data by Riedel et al.~\shortcite{riedel10}, which returned no extractions in our case.}

 {\bf SUP} is a supervised system which learns a log-linear model using the set of features for relation extraction proposed by Mintz et al.~\shortcite{mintz09}. The supervision is provided by four annotators hired on \url{odesk.com} who rated themselves as experts for data entry, and were encouraged to use any tool of their choice for annotation. Each annotator was asked to spend 1 hour per relation to identify sentences in the development corpus containing that relation and marking its arguments. 
To control the variation related to the order in which relations were presented (users get faster with time),
we used a Latin square design and paid for 1 additional hour before the experiment to allow users to get familiar with the task. Negative examples were added as in the distantly supervised cases.

\paragraph{Datasets} We used the New York Times Annotated
Corpus~\cite{sandhaus2008} comprising 1.8M news articles (45M sentences) published
between 1987 and 2007. A random half of the articles were used for development, the other
half for testing. 

\paragraph{Relations} We selected four relations: {\sf attendedSchool (person,school)}, {\sf founded (founder,organization)}, {\sf killed (killer,victim)}, and {\sf married (spouse1,spouse2)}.
These relations were selected because they cover a range of domains, they were part
in previous evaluations~\cite{ji11,doddington04,roth04}, and they do not 
require recognition of uncommon entity types.\footnote{\instaread's bootstrap rule induction and
core linguistic rules currently only target binary relations, but not entity types. To identify named
entities of types {\sf person} and {\sf organization}, we thus used the Stanford NER system. To handle
relation {\sf attendedSchool} we additionally created a 
recognizer for type {\sf school} by listing 30 common head words such as
`University' before the experiment. This process took under 5 minutes.} For preprocessing, we used the CJ Parser~\cite{CharniakJ05} and Stanford's dependency~\cite{marneffe06}, coreference~\cite{RaghunathanLRCSJM10}, and NER~\cite{finkel-acl05} systems.

\paragraph{Metrics} Extractions were counted on a mention level, which means that an
extraction consisted of both a pair of strings representing named entities as well as a reference to the sentence expressing the relation.  To measure
precision, we sampled 100 extractions and
manually created annotations following the ACE guidelines~\cite{doddington04}.

\subsection{Comparing InstaRead to Baselines}

Overall results are summarized in Table~\ref{tab:extrnyt}. In the case of \instaread, precision was 90\% or higher\footnote{The high precision in the case of \instaread\ may seem surprising, but is in fact easy to attain for many relations. Since every change the user makes to the rule set immediately triggers a re-evaluation and visual presentation of extractions and their sentences, the user can quickly adapt the rule set until she is satisfied with precision on the training set.
There is generally little overfitting, due to the training set being large and the rules not being automatically selected but created by a human with intuitions about language.

This contrasts with SUP, where a fixed feature set leads to high precision on one relation ({\sf founded}), but low precision on another ({\sf attendedSchool}). Without interactive feedback, it is very challenging to create an effective feature set as well as create effective annotated examples, especially negative ones. 
} for each of the four relations, and each extractor returned thousands of tuples. 
For the three baselines, results varied between relations but in all cases significantly fewer extractions were
returned, and in all but two cases precision was significantly lower. 
The most challenging of the four relations was {\em killed}, since it can be expressed in many
different ways, and many such expressions have multiple meanings. At the same time, mentions of the {\em killed} relation occur less frequently in the corpus than mentions of the other three relations.
The supervised baseline did not return results,
and the distantly supervised systems could not be applied because Freebase did not contain instances
for the {\em killed} relation.

\setlength{\tabcolsep}{1ex}
\begin{table}[tb]
\centering
\begin{tabular}{lrrrr}
  \multicolumn{2}{r}{\footnotesize{\sf attendedSchool}} & \footnotesize{\sf founded} & \footnotesize{\sf killed } & \footnotesize{\sf married } \\
\hline
InstaRead (rules) &     94        &  97    &  141   &  48  \\
SUP (examples)          &    68         & 79    &  36   & 52 \\
\hline
\end{tabular}
\caption{Manual input generated in one hour of time. In the supervised
case, annotators had difficulty finding examples for the {\em killed} relation
which had fewer mentions in the corpus. In contrast, \instaread's effort-reducing
features, such as rule suggestions, made it easy to find examples and add relevant 
rules quickly. Our user of \instaread\ actually generated more rules for this relation,
in the allotted time, due to the larger number of syntactic variations.
}
\label{tab:feedback}
\end{table}

\begin{figure*}[ht]
\centering
\includegraphics[trim={0 0 0 0},clip,width=3in]{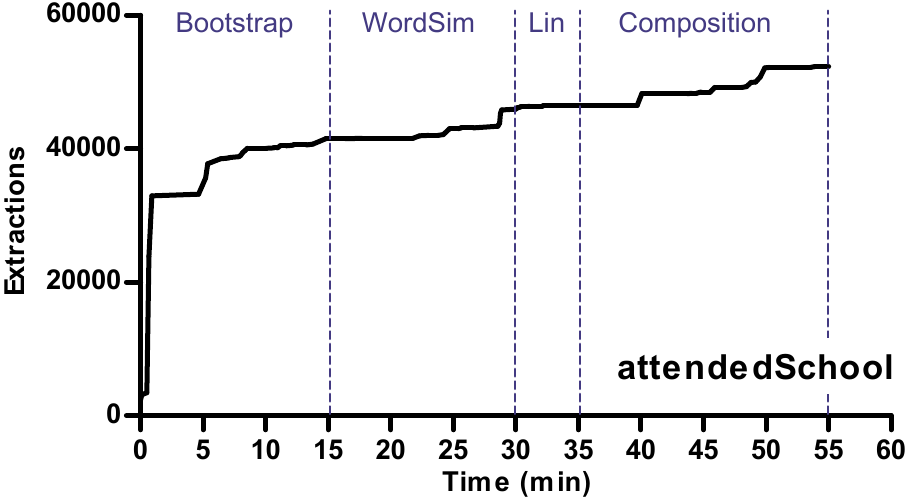}
\hspace{0.3in}
\includegraphics[trim={0 0 0 0},clip,width=3in]{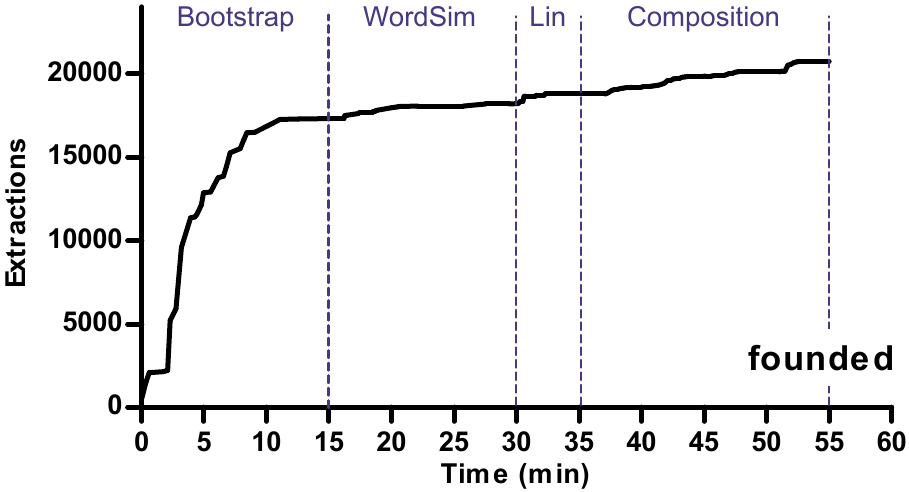}
\\
\vspace{0.3in}
\includegraphics[trim={0 0 0 0},clip,width=3in]{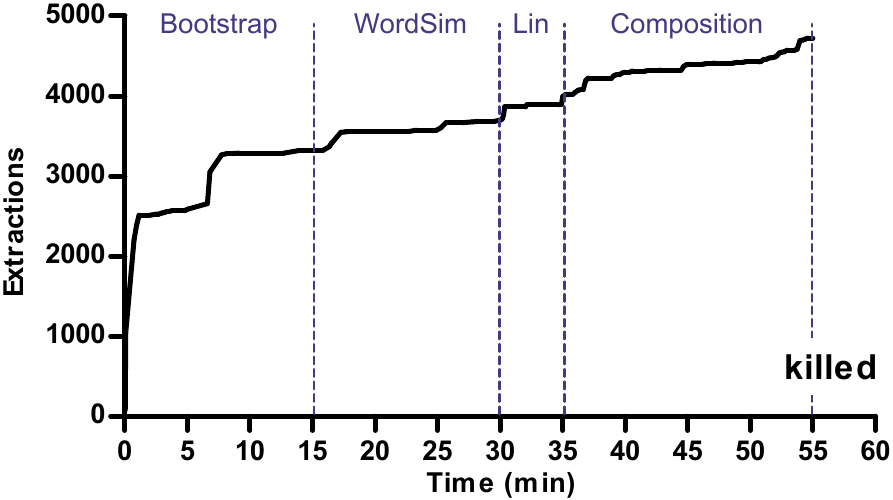}
\hspace{0.3in}
\includegraphics[trim={0 0 0 0},clip,width=3in]{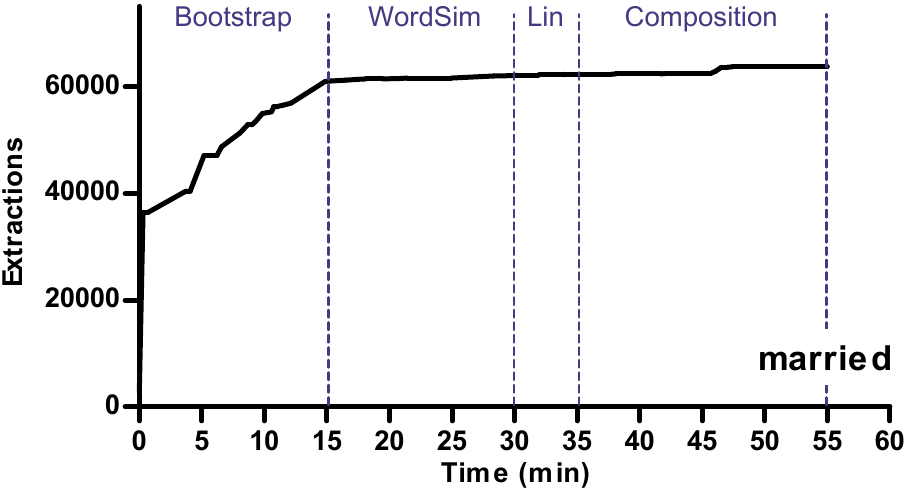}
\caption{Number of extractions on an independent test set while using \instaread\ for 55 minutes. {\em Bootstrap} (Section~\ref{sec:accelerators}) captures a large number of extractions quickly, but does not yield additional gains after a few minutes. {\em WordSim} (Section~\ref{sec:accelerators}) enables slow, but consistent gains. {\em Linguistics} (Section~\ref{sec:accelerators}) provides a small gain. {\em Composition} (Section~\ref{sec:rulelanguage}) is helpful when there exist a large number of lexical stems that imply a relation (e.g. for the killed relation).}
\label{fig:comparetools}
\end{figure*}

\paragraph{User Feedback} Looking more carefully at the feedback supplied by our users, we found that one hour use of \instaread\ yielded 95 rules on average. This compares to an average of 59 examples per hour annotated by users
in the supervised case. \instaread's effort-reducing features made it easy to find relevant sentences and
add rules quickly, which frequently only required confirmation of a system-generated suggestion.
Users in the supervised case had difficulty finding sentences expressing
the relations. Two of the annotators reported that they started off reading the text corpus linearly, but 
barely found any examples that way. They later searched by keywords (`College') and wildcards (`marr*').
With 77 and 89 examples per hour these users found more examples (but not necessarily more variations)
than users who scanned the corpus linearly and found 19 and 51 examples per hour.

Table~\ref{tab:feedback} shows a breakdown by relation, and reveals a striking difference between
\instaread\ and SUP for the {\em killed} relation. In the supervised case, users were able to identify
far fewer examples for this relation than others. In contrast, our user of \instaread\ actually generated most rules for this relation. 
This shows that \instaread\ did not suffer from the problem of finding examples.
In fact, as we will see, \instaread's effort-reducing features were actually most effective for this relation, and
the larger number of rules was necessary to cover a larger set of variations.

\paragraph{Impact of Effort-Reducing Features} Figure~\ref{fig:comparetools} shows the contribution of each of the four features on number of extractions. The vast majority of extractions, 84\%, were obtained by rules created during the {\em Bootstrap} phase.  {\em Bootstrap} has the ability to aggregate over many potential rules and then rank those taking into account the number of extractions. This ranking ensures that user effort is directed to rules which are likely to matter most. Such ranking is not possible with the {\em WordSim} feature, which, however, has a different advantage: It can find rarely used ways of expressing a relation. In contrast, {\em Bootstrap} only works if the same relation instance is expressed multiple times in different ways. We therefore often observe that it provides no more improvement after a few minutes of use. 3.4\% of extractions were obtained by rules created during the {\em WordSim} phase, 2.6\% during the {\em Linguistics} phase, and 9.5\% during the {\em Composition} phase. 

Figure~\ref{fig:comparetools} further reveals differences between relations. For {\em married}, the relatively
small number of common variations were already captured in only 15 minutes, after which {\em WordSim}, {\em Linguistics}, and {\em Composition} features provided little benefit. For {\em killed}, however, each of the
four effort-reducing features substantially increased the number of extractions. This shows that \instaread's ensemble of effort-reducing features was effective in guiding our user to the many variations of the {\em killed} relation.

\paragraph{Analysis of \instaread's Errors} \instaread's precision errors for the four relations were to a large degree caused by errors in preprocessing, 
especially dependency extraction (55\%) and NER (24\%). Only 21\% of precision errors were caused by overly general rules that the expert
user had developed. All were due to ambiguities of the words {\sf fell}, {\sf executor}, and {\sf built}.
While the effort-reducing features have been designed to increase recall, \instaread's focus on only deterministic rules is not adequate to easily handle such ambiguities -- a shortcoming we would like
to address in future work.

\paragraph{Enhancing Supervised Extraction} Finally, we are interested in knowing if an increase in time would let users in the supervised case
match \instaread's results. We therefore combined the annotations of all four annotators; each relation's
examples thus corresponded to four hours of manual effort. 
Trained on this data, SUP returned more extractions (attendedSchool -- 51,492, founded -- 7,482, killed -- 220, married -- 24,866), but precision remained low and in two cases even decreased slightly (attendedSchool -- 12\%, founded -- 97\%, killed -- 0\%, married -- 34\%). In summary, additional time {\em does} improve performance, but many more hours of annotation effort would be required to reach performance comparable to \instaread.

The features we selected have been shown to work well for many relations~\cite{mintz09}, but it is still possible that better features could improve the supervised learning algorithm's performance. However, feature engineering itself takes considerable effort, usually measured in weeks, which would defeat our goal of building complete extractors quickly. It will be an important area for future work to determine if \instaread\ can be adapted to support rapid authoring of rules that define feature templates, perhaps providing even better overall performance on a limited engineering budget.

\paragraph{Comparing to Extreme Extraction Work} It is impossible to compare directly to Freedman et al~\cite{freedman11}, since we were unable to acquire their datasets. While their approach yielded an average precision of 53\% across 5 relations, they used 50 hours of manual engineering and furthermore those hours were spread across several different experts, each with knowledge of a specific tool. 

Unlike \instaread's {\em Bootstrap} feature, their bootstrap learner ran autonomously without user
interaction, but contributed little to increase overall performance. We suspect that \instaread's user in the loop, instant execution, integration of coreference information, and larger corpus contributed to perceived differences in effectiveness.
Section~\ref{sec:instarelated} discusses further differences and similarities of Freedman et al's work and \instaread.

\subsection{Real-world Use By Engineers}

Our experiments so far tested \instaread's effectiveness for a trained
expert; in our final experiment, we evaluated if the system was also usable by
engineers without NLP background.

We recruited four senior undergraduate students in Computer Science who used
\instaread\ as part of a quarterly class project to develop 30 relation extractors
for the TAC-KBP slot filling challenge. In six meetings, usage of the tool was 
explained and qualitative feedback collected.

All four subjects were able to use the system with little instruction, all were able to
develop extractors, and all four subjects reported that the tool made it easier for them 
than if they had to write their own code.
Among the 27 extractors that were created, median precision was 94\% (mean 75\%), 
and median number of extractions on NYTimes data was 2283 (mean 8741).
For two relations, no extractor was created due to the difficulty in creating custom
entity type recognizers, and for one relation due to an implementation error.
Mean precision was negatively affected by six relations which required custom
entity type recognizers. \instaread\ currently has no support for developing 
entity type recognizers, a shortcoming which we would like to address in future work. 
Another important area for improvement is the interface to manage sets of rules.
The subjects found it was often easier for them to manage rule sets in code 
(as strings of logical expression), because they could add their own comments, 
re-arrange, and keep track of multiple versions.

%% file: discussion.tex
\section{Conclusions and Future Work}
\label{sec:instadiscuss}

Many successful applications of IE rely on large amounts of manual
engineering, which often requires the laborious selection of rules
to be used as extraction patterns or features. 

This paper presents ways to streamline this process,
proposing an ensemble of methods that enable three properties: an expressive
rule language, guidance that leads users to promising rules, and instant
rule testing. Our experiments demonstrate that \instaread\ enables
experts to develop quality relation extractors in under one hour --
an order of magnitude reduction in effort from Freedman et al.~\shortcite{freedman11}. 
To stimulate continued progress in the area, we release our data as explained in footnote 1. 

The experiments also point to two promising directions to further reduce
manual effort:


\paragraph{Richness of Interactions}
With the {\em Bootstrap}, {\em WordSim}, {\em Linguistics}, and {\em Composition}
features, \instaread\ offered a variety of interactions, all of which contributed to increased
recall while maintaining high-precision.
{\em Bootstrap} was particularly effective, but did not allow further improvements after a few
minutes of use. {\em WordSim} did not show this problem, but expanded recall more slowly.
{\em Composition} was very effective for some relations. {\em Linguistics} yielded smaller gains,
but required less effort. Future improvements to
cover additional syntactic variations, such as participle phrases, may increase gains.

We consider such variety of interactions essential, and thus plan to include interactions
for clustering phrases, providing databases of instances for distant supervision,
editing ontologies, providing validative feedback, and annotating sentences.
Determining the relative importance of such interactions will be an important future challenge.

\paragraph{Deep Integration of Algorithms}

Perhaps even greater potential, however, may lie in more tightly integrating \instaread's components. 
Our analysis of precision errors revealed that the majority of precision errors were caused by inaccurate preprocessing, and we believe that jointly taking into account manually created rules as well as the $k$ best outputs of the preprocessing components could improve results.
We further suspect learning-based techniques may be
particularly important for tasks such as NER, where there exist many ambiguities, while
rule-based techniques may work well for tasks such as defining implicature between phrases.

\instaread's {\em Boostrap} feature could also be improved. It currently 
already leverages coreference clusters and syntactic dependencies. In fact, coreference information which greatly increases recall may explain much of bootstrap learning's observed high effectiveness compared
to Freedman et al.'s work. In the future, we would like to enable {\em Bootstrap} to also take into 
account our core linguistic rules and the ability to decompose rules. Such integration may expand recall,
and interestingly, might also simplify the interaction with the user. Since the integrated components
enable rules with higher coverage, fewer, more distinct rules would be returned.

%% file: appendix.tex
\section*{Appendix}
\label{sec:translation}

\begin{figure}[ht]
\centering
\setlength\parindent{24pt}
\begin{minipage}{\linewidth}
$\sf r(t) \Leftarrow str2span(\textrm`Lee\,Harvey\,Oswald\textrm', s) \wedge$\\
\phantom{$\sf r(t) \Leftarrow $}$\sf span2pos(s,p) \wedge nsubj(c,p) \wedge$\\
\phantom{$\sf r(t) \Leftarrow $}$\sf token(c,t)$
\end{minipage}
\\
\begin{minipage}{\linewidth}
\centering
$\vardarrow{0.4cm}$translation
\end{minipage}
\\
\vspace{.1in}
\setlength\parindent{50pt}
\begin{minipage}{\linewidth}
\begin{spacing}{1.2}
\tiny{
\begin{verbatim}
SELECT ti4.tokenID
FROM tokenInst ti0, tokenInst ti1,
  tokenInst ti2, tokenInst ti3,
  dependencyInst di0, tokenInst ti4
WHERE ti4.offset = di0.from
  AND ti4.sentenceID = di0.sentenceID
  AND di0.to = ti3.offset
  AND di0.sentenceID = ti3.sentenceID
  AND di0.dependencyID = 11
  AND ti3.offset < ti2.offset + 1
  AND ti3.offset >= ti0.offset
  AND ti3.sentenceID = ti0.sentenceID
  AND ti2.tokenID = 79216
  AND ti1.tokenID = 6058
  AND ti0.tokenID = 5322
  AND ti0.offset + 2 = ti2.offset
  AND ti0.sentenceID = ti2.sentenceID
  AND ti0.offset + 1 = ti1.offset
  AND ti0.sentenceID = ti1.sentenceID\end{verbatim}
\vspace{-.15in}
}
\end{spacing}
\end{minipage}
\caption{Translation of a (safe) expression in first-order logic to SQL. The expression
returns verbs for which Lee Harvey Oswald appears as subject. $\tt str2spans$ and $\tt span2pos$
are intensional predicates, $\tt nsubj$ and $\tt token$ are extensional.
Each predicate gets translated into a fragment of SQL; the fragments are then combined into
a single SQL query, which can be efficiently executed.
}
\label{fig:instaintext}
\end{figure}

\begin{figure}[ht]
\begin{minipage}{\linewidth}
\tiny{
$\sf \killOfNom(c,b) \Leftarrow \ldots$ \\
$\sf \ldots \prepof(c,b) \wedge \token(c, \textrm`assassination\textrm')$ \\
$\sf \ldots \prepof(c,b) \wedge \token(c, \textrm`execution\textrm')$ \\
$\sf \ldots \prepof(c,b) \wedge \token(c, \textrm`felling\textrm')$ \\
$\sf \ldots \prepof(c,b) \wedge \token(c, \textrm`killing\textrm')$ \\
$\sf \ldots \prepof(c,b) \wedge \token(c, \textrm`shooting\textrm')$ \\
$\sf \ldots \prepof(c,b) \wedge \token(c, \textrm`slaughter\textrm')$ \\
$\sf \ldots \prepof(c,b) \wedge \token(c, \textrm`slaying\textrm')$ \\
$\sf \ldots \prepof(c,b) \wedge \token(c, \textrm`stabbing\textrm')$ \\
$\sf \ldots \prepof(c,b) \wedge \token(c,\textrm`murder\textrm')$ \\
$\sf \ldots \nn(c,b) \wedge \token(c, \textrm`assassination\textrm')$\\
$\sf \ldots \nn(c,b) \wedge \token(c,\textrm`murder\textrm')$ \\
$\sf \ldots$ \\
$\sf \killerRole(c) \Leftarrow \ldots$ \\
$\sf \ldots \token(c, \textrm`assassin\textrm')$ \\
$\sf \ldots \token(c, \textrm`murderer\textrm')$ \\
$\sf \ldots$ \\ 
$\sf \killingBNF(c,b) \Leftarrow \ldots$ \\
$\sf \ldots \dobj(c,b) \wedge \token(c, \textrm`assassinate\textrm')$ \\
$\sf \ldots \dobj(c,b) \wedge \token(c, \textrm`murder\textrm')$ \\
$\sf \ldots$ \\
$\sf \killingBInf(d,b) \Leftarrow \ldots$ \\
$\sf \ldots \dobj(d,b) \wedge \token(d, \textrm`assassinating\textrm')$ \\
$\sf \ldots \dobj(d,b) \wedge \token(d, \textrm`murdering\textrm')$ \\
$\sf \ldots$ \\
$\sf \killed(a,b) \Leftarrow \person(b) \wedge \person(a) \wedge (a \neq b) \wedge \ldots $ \\
$\sf \ldots \actInd(a,c,\textrm`confess\textrm') \wedge \prepcto(c,d) \wedge \killingBInf(d,b)$ \\
$\sf \ldots \actInd(a,c,\textrm`confess\textrm') \wedge \prepto(c,d) \wedge \killOfNom(d,b)$ \\
$\sf \ldots \actInd(a,c,\textrm`assassinate\textrm') \wedge \dobj(c,b)$ \\
$\sf \ldots \actInd(a,c,\textrm`murder\textrm') \wedge \dobj(c,b)$ \\
$\sf \ldots$ \\
$\sf \ldots \agent(c,a) \wedge \partmod(b,c) \wedge \token(c,\textrm`assassinated\textrm')$ \\
$\sf \ldots \agent(c,a) \wedge \partmod(b,c) \wedge \token(c,\textrm`murdered\textrm')$ \\
$\sf \ldots$ \\
$\sf \ldots \agent(c,a) \wedge \rcmod(b,c) \wedge \token(c,\textrm`assassinated\textrm')$ \\
$\sf \ldots \agent(c,a) \wedge \rcmod(b,c) \wedge \token(c,\textrm`murdered\textrm')$ \\
$\sf \ldots$ \\
$\sf \ldots \appos(a,c) \wedge \poss(c,b) \wedge \killerRole(c)$ \\
$\sf \ldots \appos(a,c) \wedge \prepof(c,b) \wedge \killerRole(c)$ \\
$\sf \ldots \appos(c,a) \wedge \poss(c,b) \wedge \killerRole(c)$ \\
$\sf \ldots \appos(c,a) \wedge \prepof(c,b) \wedge \killerRole(c)$ \\
$\sf \ldots \dep(a,c) \wedge \dobj(c,b) \wedge \token(c,\textrm`assassinated\textrm')$ \\
$\sf \ldots \dep(a,c) \wedge \dobj(c,b) \wedge \token(c,\textrm`murdered\textrm')$ \\
$\sf \ldots \infmod(a,c) \wedge \killingBNF(c,b)$ \\
$\sf \ldots \nsubj(c,a) \wedge \prepin(c,d) \wedge \token(c,\textrm`suspect\textrm') \wedge \killOfNom(d,b)$ \\
$\sf \ldots \nsubj(c,a) \wedge \xcomp(c,d) \wedge \killingBInf(d,b)$ \\
$\sf \ldots \partmod(a,c) \wedge \killingBInf(c,b)$ \\
$\sf \ldots \partmod(a,c) \wedge \prepcfor(c,d) \wedge \token(c,\textrm`sentenced\textrm') \wedge \killingBInf(d,b)$ \\
$\sf \ldots \partmod(a,c) \wedge \prepcof(c,d) \wedge \token(c,\textrm`accused\textrm') \wedge \killingBInf(d,b)$ \\
$\sf \ldots \partmod(a,c) \wedge \prepcof(c,d) \wedge \token(c,\textrm`convicted\textrm') \wedge \killingBInf(d,b)$ \\
$\sf \ldots \partmod(a,c) \wedge \prepcwith(c,d) \wedge \token(c,\textrm`charged\textrm') \wedge \killingBInf(d,b)$ \\
$\sf \ldots \partmod(a,c) \wedge \prepin(c,d) \wedge \token(c,\textrm`wanted\textrm') \wedge \prepwith(d,e) \wedge \token(d,\textrm`connection\textrm') \wedge $ \\
\phantom{$\sf \ldots $} $\killOfNom(e,b)$ \\
$\sf \ldots \partmod(a,c) \wedge \prepto(c,d) \wedge \token(c,\textrm`linked\textrm') \wedge \killOfNom(d,b)$ \\
$\sf \ldots \passInd(a,c,\textrm`accuse\textrm') \wedge \prepof(c,d) \wedge \killOfNom(d,b)$ \\
$\sf \ldots \passInd(a,c,\textrm`charge\textrm') \wedge \prepcwith(c,d) \wedge \killingBInf(d,b)$ \\
$\sf \ldots \passInd(a,c,\textrm`charge\textrm') \wedge \prepwith(c,d) \wedge \killOfNom(d,b)$ \\
$\sf \ldots \passInd(a,c,\textrm`convict\textrm') \wedge \prepcfor(c,d) \wedge \killingBInf(d,b)$ \\
$\sf \ldots \passInd(a,c,\textrm`convict\textrm') \wedge \prepcof(c,d) \wedge \killingBInf(d,b)$ \\
$\sf \ldots \passInd(a,c,\textrm`convict\textrm') \wedge \prepcof(c,d) \wedge \prepin(d,e) \wedge \token(d,\textrm`taking\textrm') \wedge \killOfNom(e,b)$ \\
$\sf \ldots \passInd(a,c,\textrm`convict\textrm') \wedge \prepfor(c,d) \wedge \killOfNom(d,b)$ \\
$\sf \ldots \passInd(a,c,\textrm`convict\textrm') \wedge \prepin(c,d) \wedge \prepof(d,b) \wedge \token(d,\textrm`death\textrm')$ \\
$\sf \ldots \passInd(a,c,\textrm`convict\textrm') \wedge \prepof(c,d) \wedge \killOfNom(d,b)$ \\
$\sf \ldots \passInd(a,c,\textrm`link\textrm') \wedge \prepto(c,d) \wedge \killOfNom(d,b)$ \\
$\sf \ldots \passInd(a,c,\textrm`sentence\textrm') \wedge \prepfor(c,d) \wedge \killOfNom(d,b)$ \\
$\sf \ldots \passInd(a,c,\textrm`want\textrm') \wedge \prepin(c,d) \wedge \prepwith(d,e) \wedge \token(d,\textrm`connection\textrm') \wedge \killOfNom(e,b)$ \\
$\sf \ldots \passInd(b,c,\textrm`assassinate\textrm') \wedge \agent(c,a)$ \\
$\sf \ldots \passInd(b,c,\textrm`gun\textrm') \wedge \prt(c,d) \wedge \token(d,\textrm`down\textrm') \wedge \agent(c,a)$ \\
$\sf \ldots \passInd(b,c,\textrm`murder\textrm') \wedge \agent(c,a)$ \\
$\sf \ldots \passInd(l,c,\textrm`take\textrm') \wedge \token(l,\textrm`life\textrm') \wedge \poss(l,b) \wedge \agent(c,a)$ \\
$\sf \ldots \passInd(l,c,\textrm`take\textrm') \wedge \token(l,\textrm`life\textrm') \wedge \prepof(l,b) \wedge \agent(c,a)$ \\
$\sf \ldots$ \\
$\sf \ldots \poss(c,a) \wedge \killOfNom(c,b)$ \\
$\sf \ldots \poss(c,a) \wedge \nsubjpass(d,c) \wedge \token(c,\textrm`name\textrm') \wedge \prepto(d,e) \wedge \token(d,\textrm`linked\textrm') \wedge \killOfNom(e,b)$ 
$\sf \ldots \prepby(c,a) \wedge \killOfNom(c,b)$ \\
$\sf \ldots \rcmod(a,c) \wedge \dobj(c,b) \wedge \token(c,\textrm`assassinated\textrm')$ \\
$\sf \ldots \rcmod(a,c) \wedge \dobj(c,b) \wedge \token(c,\textrm`murdered\textrm')$ \\
$\sf \ldots$ \\
$\sf \ldots \token(a,\textrm`suspect\textrm') \wedge \nsubj(c,a) \wedge \prepin(c,d) \wedge \prepof(d,b) \wedge \nn(e,d) \wedge \token(e,\textrm`murder\textrm') \wedge$ \\
\phantom{$\sf \ldots $} $\token(d,\textrm`trial\textrm')$ \\
$\sf \ldots \xsubj(c,a) \wedge \killingBNF(c,b)$ \\
$\sf \ldots$ \\
$\sf \ldots$ \\
}
\end{minipage}
\caption{Selected extraction rules created for relation $\sc killed$ during the experiment. Many extractions were obtained during the {\em Bootstrap} phase, which suggested rules combining syntactic dependencies (eg. $\nn$) and lexical information (eg. $\token$). Users selected from these suggestions, but also adapted them by adding constraints (eg. $\prt$, $\neq$, $\person$). {\em WordSim} added lexical variety (eg. $\killOfNom$), and {\em Linguistics} covered additional verb inflections (encoded by predicates $\actInd$ and $\passInd$). {\em Composition} introduced re-usable components (eg. $\killerRole$, $\killingBNF$).
}
\label{fig:exampleruleset}
\end{figure}

%% file: main.bbl
\begin{thebibliography}{10}

\bibitem{appelt98}
D.~E. Appelt and B.~Onyshkevych.
\newblock The common pattern specification language.
\newblock In {\em Proceedings of a Workshop Held at TIPSTER 98}, pages 23--30.
  Association for Computational Linguistics, 1998.

\bibitem{bollacker08}
K.~D. Bollacker, C.~Evans, P.~Paritosh, T.~Sturge, and J.~Taylor.
\newblock Freebase: a collaboratively created graph database for structuring
  human knowledge.
\newblock In {\em SIGMOD Conference}, pages 1247--1250, 2008.

\bibitem{carlson09b}
A.~Carlson, J.~Betteridge, E.~R.~H. Jr., and T.~M. Mitchell.
\newblock Coupling semi-supervised learning of categories and relations.
\newblock In {\em Proceedings of the NAACL HLT Workskop on Semi-supervised
  Learning for Natural Language Processing}, 2009.

\bibitem{tokensregex2014}
A.~X. Chang and C.~D. Manning.
\newblock {TokensRegex}: Defining cascaded regular expressions over tokens.
\newblock Technical Report CSTR 2014-02, Department of Computer Science,
  Stanford University, 2014.

\bibitem{CharniakJ05}
E.~Charniak and M.~Johnson.
\newblock Coarse-to-fine n-best parsing and maxent discriminative reranking.
\newblock In {\em Proceedings of the Annual Meeting of the Association for
  Computational Linguistics (ACL)}, 2005.

\bibitem{ChiticariuKLRRV10}
L.~Chiticariu, R.~Krishnamurthy, Y.~Li, S.~Raghavan, F.~Reiss, and
  S.~Vaithyanathan.
\newblock Systemt: An algebraic approach to declarative information extraction.
\newblock In {\em Proceedings of the Annual Meetings of the Association for
  Computational Linguistics (ACL)}, pages 128--137, 2010.

\bibitem{doddington04}
G.~Doddington, A.~Mitchell, M.~Przybocki, L.~Ramshaw, S.~Strassel, and
  R.~Weischedel.
\newblock Ace program - task definitions and performance measures.
\newblock In {\em Proceedings of the International Conference on Language
  Resources and Evaluation (LREC)}, pages 837--840, 2004.

\bibitem{2009Domingos}
P.~Domingos and D.~Lowd.
\newblock {\em Markov Logic: An Interface Layer for Artificial Intelligence}.
\newblock Synthesis Lectures on Artificial Intelligence and Machine Learning.
  Morgan {\&} Claypool Publishers, 2009.

\bibitem{druck09}
G.~Druck, B.~Settles, and A.~McCallum.
\newblock Active learning by labeling features.
\newblock In {\em Proceedings of the Conference on Empirical Methods in Natural
  Language Processing (EMNLP)}, pages 81--90, 2009.

\bibitem{finkel-acl05}
J.~R. Finkel, T.~Grenager, and C.~Manning.
\newblock Incorporating non-local information into information extraction
  systems by gibbs sampling.
\newblock In {\em Proceedings of the Annual Meeting of the Association for
  Computational Linguistics (ACL)}, pages 363--370, 2005.

\bibitem{freedman11}
M.~Freedman, L.~A. Ramshaw, E.~Boschee, R.~Gabbard, G.~Kratkiewicz, N.~Ward,
  and R.~M. Weischedel.
\newblock Extreme extraction - machine reading in a week.
\newblock In {\em Proceedings of the Conference on Empirical Methods in Natural
  Language Processing (EMNLP)}, pages 1437--1446, 2011.

\bibitem{gabbard11}
R.~Gabbard, M.~Freedman, and R.~M. Weischedel.
\newblock Coreference for learning to extract relations: Yes virginia,
  coreference matters.
\newblock In {\em Proceedings of the Annual Meeting of the Association for
  Computation Linguistics (ACL)}, pages 288--293, 2011.

\bibitem{ganchev10}
K.~Ganchev, J.~Gra\c{c}a, J.~Gillenwater, and B.~Taskar.
\newblock Posterior regularization for structured latent variable models.
\newblock {\em Journal of Machine Learning Research}, 11:2001--2049, 2010.

\bibitem{hoffmann11}
R.~Hoffmann, C.~Zhang, X.~Ling, L.~S. Zettlemoyer, and D.~S. Weld.
\newblock Knowledge-based weak supervision for information extraction of
  overlapping relations.
\newblock In {\em Proceedings of the Annual Meeting of the Association for
  Computational Linguistics (ACL)}, pages 541--550, 2011.

\bibitem{hoffmann10}
R.~Hoffmann, C.~Zhang, and D.~S. Weld.
\newblock Learning 5000 relational extractors.
\newblock In {\em Proceedings of the Annual Meeting of the Association for
  Computational Linguistics (ACL)}, pages 286--295, 2010.

\bibitem{ji11}
H.~Ji, R.~Grishman, and H.~T. Dang.
\newblock An overview of the tac2011 knowledge base population track.
\newblock In {\em Proceedings of the Text Analysis Conference (TAC)}, 2011.

\bibitem{KrishnamurthyLRRVZ08}
R.~Krishnamurthy, Y.~Li, S.~Raghavan, F.~Reiss, S.~Vaithyanathan, and H.~Zhu.
\newblock Systemt: a system for declarative information extraction.
\newblock {\em SIGMOD Record}, 37(4):7--13, 2008.

\bibitem{lin01}
D.~Lin and P.~Pantel.
\newblock Dirt - discovery of inference rules from text.
\newblock In {\em Proceedings of the ACM SIGKDD International Conference on
  Knowledge Discovery and Data Mining (KDD)}, pages 323--328, 2001.

\bibitem{marneffe06}
M.-C.~D. Marneffe, B.~Maccartney, and C.~D. Manning.
\newblock Generating typed dependency parses from phrase structure parses.
\newblock In {\em Proceedings of the International Conference on Language
  Resources and Evaluation (LREC)}, 2006.

\bibitem{miller04}
S.~Miller, J.~Guinness, and A.~Zamanian.
\newblock Name tagging with word clusters and discriminative training.
\newblock In {\em Proceedings of the Human Language Technology Conference of
  the North American Chapter of the Association for Computational Linguistics
  (HLT-NAACL)}, 2004.

\bibitem{min12}
B.~Min, X.~Li, R.~Grishman, and A.~Sun.
\newblock New york university 2012 system for kbp slot filling.
\newblock In {\em Proceedings of the Text Analysis Conference (TAC)}, 2012.

\bibitem{mintz09}
M.~Mintz, S.~Bills, R.~Snow, and D.~Jurafsky.
\newblock Distant supervision for relation extraction without labeled data.
\newblock In {\em Proceedings of the Annual Meeting of the Association for
  Computational Linguistics (ACL)}, pages 1003--1011, 2009.

\bibitem{nakashole11}
N.~Nakashole, M.~Theobald, and G.~Weikum.
\newblock Scalable knowledge harvesting with high precision and high recall.
\newblock In {\em Proceedings of the International Conference on Web Search and
  Data Mining (WSDM)}, pages 227--236, 2011.

\bibitem{nakashole12}
N.~Nakashole, G.~Weikum, and F.~M. Suchanek.
\newblock Discovering and exploring relations on the web.
\newblock {\em {The Proceedings of the VLDB Endowment (PVLDB)}},
  5(12):1982--1985, 2012.

\bibitem{nakashole13}
N.~Nakashole, G.~Weikum, and F.~M. Suchanek.
\newblock Discovering semantic relations from the web and organizing them with
  {PATTY}.
\newblock {\em {SIGMOD} Record}, 42(2):29--34, 2013.

\bibitem{niu12}
F.~Niu, C.~Zhang, C.~Re, and J.~W. Shavlik.
\newblock Deepdive: Web-scale knowledge-base construction using statistical
  learning and inference.
\newblock In {\em In Proceedings of the Second International Workshop on
  Searching and Integrating New Web Data Sources (VLDS)}, pages 25--28, 2012.

\bibitem{niu12x}
F.~Niu, C.~Zhang, C.~R{\'{e}}, and J.~W. Shavlik.
\newblock Elementary: Large-scale knowledge-base construction via machine
  learning and statistical inference.
\newblock {\em International Journal on Semantic Web and Information Systems},
  8(3):42--73, 2012.

\bibitem{poon07}
H.~Poon and P.~Domingos.
\newblock Joint inference in information extraction.
\newblock In {\em Proceedings of the AAAI Conference on Artificial
  Intelligence}, pages 913--918, 2007.

\bibitem{RaghunathanLRCSJM10}
K.~Raghunathan, H.~Lee, S.~Rangarajan, N.~Chambers, M.~Surdeanu, D.~Jurafsky,
  and C.~D. Manning.
\newblock A multi-pass sieve for coreference resolution.
\newblock In {\em Proceedings of the Conference on Empirical Methods in Natural
  Language Processing (EMNLP)}, pages 492--501, 2010.

\bibitem{re14}
C.~R{\'{e}}, A.~A. Sadeghian, Z.~Shan, J.~Shin, F.~Wang, S.~Wu, and C.~Zhang.
\newblock Feature engineering for knowledge base construction.
\newblock {\em {IEEE} Data Engineering Bulletin}, 37(3):26--40, 2014.

\bibitem{riedel10}
S.~Riedel, L.~Yao, and A.~McCallum.
\newblock Modeling relations and their mentions without labeled text.
\newblock In {\em Proceedings of the European Conference on Machine Learning
  and Principles and Practice of Knowledge Discovery in Databases (ECML/PKDD)},
  pages 148--163, 2010.

\bibitem{Riloff96}
E.~Riloff.
\newblock Automatically generating extraction patterns from untagged text.
\newblock In {\em AAAI/IAAI, Vol. 2}, pages 1044--1049, 1996.

\bibitem{roth04}
D.~Roth and W.-T. Yih.
\newblock {A Linear Programming Formulation for Global Inference in Natural
  Language Tasks}.
\newblock In {\em Proceedings of the 2004 Conference on Computational Natural
  Language Learning (CoNLL)}, pages 1--8, 2004.

\bibitem{sandhaus2008}
E.~Sandhaus.
\newblock {\em The New York Times Annotated Corpus}.
\newblock Linguistic Data Consortium, 2008.

\bibitem{suchanek08}
F.~M. Suchanek, G.~Kasneci, and G.~Weikum.
\newblock Yago: A large ontology from wikipedia and wordnet.
\newblock {\em Elsevier Journal of Web Semantics}, 6(3):203--217, 2008.

\bibitem{sun11}
A.~Sun, R.~Grishman, W.~Xu, and B.~Min.
\newblock New york university 2011 system for kbp slot filling.
\newblock In {\em Proceedings of the Text Analysis Conference (TAC)}, 2011.

\bibitem{mihai12}
M.~Surdeanu, J.~Tibshirani, R.~Nallapati, and C.~Manning.
\newblock Multi-instance multi-label learning for relation extraction.
\newblock In {\em Proceedings of the Conference on Empirical Methods in Natural
  Language Processing (EMNLP)}, 2012.

\bibitem{thompson99}
C.~A. Thompson, M.~E. Califf, and R.~J. Mooney.
\newblock Active learning for natural language parsing and information
  extraction.
\newblock In {\em Proceedings of the International Conference on Machine
  Learning (ICML)}, pages 406--414, 1999.

\bibitem{Ullman88}
J.~D. Ullman.
\newblock {\em Principles of Database and Knowledge-Base Systems, Volume I}.
\newblock Computer Science Press, 1988.

\end{thebibliography}
